\documentclass{article}

\usepackage[final]{neurips_2022}

\usepackage[utf8]{inputenc} 
\usepackage[T1]{fontenc}    
\usepackage{hyperref}       
\usepackage{url}            
\usepackage{booktabs}       
\usepackage{amsfonts}       
\usepackage{nicefrac}       
\usepackage{microtype}      
\usepackage{xcolor}         
\usepackage[ruled,vlined]{algorithm2e}
\usepackage{tikz,booktabs,multirow,enumitem,bm}
\usepackage{colortbl}
\usepackage{tabularx}
\usepackage{amsmath,amsthm,amsfonts}

\newtheorem{proposition}{Proposition}

\newcommand{\bx}{\mathbf{x}}
\newcommand{\by}{\mathbf{y}}
\newcommand{\bs}{\mathbf{s}}
\newcommand{\bi}{\mathbf{i}}

\newcommand{\one}[0]{\textbf{1}}
\newcommand{\zero}[0]{\textbf{0}}

\title{A Reduction to Binary Approach for Debiasing Multiclass Datasets}

\author{%
  Ibrahim~Alabdulmohsin \\
  Google Research\\
  Z\"urich, Switzerland \\
  \texttt{ibomohsin@google.com} \\
   \And
   Jessica Schrouff \\
  Google Research\\
   London, United Kingdom \\
   \texttt{schrouff@google.com} \\
   \AND
   Oluwasanmi Koyejo \\
  Google Research\\
   Mountain View, United States \\
   \texttt{sanmik@google.com} \\
}

\begin{document}

\maketitle

\begin{abstract}

We propose a novel reduction-to-binary (R2B) approach that enforces demographic parity for multiclass classification with non-binary sensitive attributes via a reduction to a sequence of binary debiasing tasks. We prove that R2B satisfies optimality and bias guarantees and demonstrate empirically that it can lead to an improvement over two baselines: (1) treating multiclass problems  as multi-label by debiasing labels independently and (2) transforming the features instead of the labels. Surprisingly, we also demonstrate that independent label debiasing yields competitive results in most (but not all) settings. We validate these conclusions on synthetic and real-world datasets from social science, computer vision, and healthcare. 
\end{abstract}

\section{Introduction}
Several studies have demonstrated that predictors are susceptible to unintended bias -- for example, deep neural networks (DNNs) can amplify spurious correlations in the training data \citep{hendricks2018women,bolukbasi2016man,caliskan2017semantics,zhao2017men,yang2020towards,wang2020towards,stock2018convnets}.  Moreover, error disparities can arise, where the performance of the model for minority groups is disproportionately worse than for the majority \citep{zhao2017men,buolamwini2018gender,deuschel2020uncovering}. Studies, such as \citet{buolamwini2018gender} and \citet{wang2020revise}, observe that one source of this disparity is that datasets may reflect societal stereotypes, thereby highlighting the importance of debiasing datasets. 

Nevertheless, despite the proliferation of research on algorithmic fairness in recent years, very few methods exist that can handle multiclass classification tasks with non-binary sensitive attributes. This gap is particularly noteworthy given that this setting is the norm in real-world applications, not the exception. Considering multiclass classification, out of the 70+ image classification datasets in the  TensorFlow Dataset catalog \citep{tensorflow2015-whitepaper}, less than 10\% are for binary classification problems. Along similar lines, considering non-binary sensitive attributes, the seven protected attributes according to the US Equal Credit Opportunity Act of 1974 \citep{usequalcredit1974} are non-binary, such as gender, race, religion, and age.

\begin{figure}
    \footnotesize
    \centering
    \begin{tabular}{lc}
        $s = \begin{bmatrix}0\\1\\0\end{bmatrix}$,\quad\quad&
         $Y = \begin{bmatrix}\frac{1}{2} & \frac{1}{2} & 0\\
         1 & 0 & 0\\
         0 & 0 & 1
         \end{bmatrix}\quad \overset{\text{debias}}{\longrightarrow}\quad
         \hat Y = \begin{bmatrix}1 & \frac{1}{2} & 0\\
         \frac{1}{2} & \frac{1}{4} & \frac{1}{2}\\
         0 & 0 & 1
         \end{bmatrix}
         \quad \overset{\text{normalize}}{\longrightarrow}\quad
         \tilde Y = \begin{bmatrix}\frac{2}{3} & \frac{1}{3} & 0\\
         \frac{2}{5} & \frac{1}{5} & \frac{2}{5}\\
         0 & 0 & 1
         \end{bmatrix}
         $
    \end{tabular}
    \caption{An illustration of why treating multiclass problems as multi-label may not achieve demographic parity (DP). We assume a binary sensitive attribute $s$ and a matrix $Y$ of label conditional probability consisting of three data records (rows) and three classes (columns). Multiclass DP exists if $\mathbb{E}[\by|\bs=0] \neq \mathbb{E}[\by|\bs=1]$, where $\by\in\mathbb{R}^3$ a vector of probability scores over the three possible classes (see Equation \ref{eq:multiclass:dp}). Observe that $Y$ does not satisfy DP because the probability scores are not independent of $\bs$. The matrix $\hat Y$ (in middle) debiases every label separately (to see this, average $s=0$ rows and compare to the $\bs=1$ row). But, to construct proper multiclass scores, its rows are normalized into probability distributions in $\tilde Y$, which reintroduces bias. See Appendix \ref{appendix:ml} for details.}
    \label{fig:multilabel_fails}
\end{figure}

Two approaches are available to handle such a broad setting, to the best of our knowledge. One option is to view  the multiclass problem as  multi-label and debias every label separately, i.e. transform the labels such that they are uncorrelated with the sensitive attribute, before normalizing the output into a valid probability distribution. However, this approach lacks fairness guarantees since normalizing the output can re-introduce bias (see Figure \ref{fig:multilabel_fails} for a cartoon illustration). Second, one can debias the instance \emph{features} instead of the labels irrespective of the number of classes, such as using the demographic parity remover \citep{feldman2015certifying}, which debiases every feature separately in a rank-preserving manner. However, debiasing the features can be suboptimal when the number of features is large as we demonstrate in Section \ref{sect::exp}. Note, in particular, that the latter approach requires a good estimate of the true cumulative distribution function (CDF) for each feature, but the uniform rate of convergence of the multivariate empirical CDF to the true CDF decreases as the number of features grows \citep{naaman2021tight}. 
We explore the performance of this approach and illustrate in Appendix \ref{appendix:dpremover} how it can fail when using, for example, equal-width binning to estimate the CDFs.

To address this gap in the literature, we propose a reduction-to-binary (R2B) approach for debiasing multiclass datasets that can accommodate an arbitrary number of classes and groups and does not require access to the sensitive attribute at prediction time. The proposed algorithm is based on the alternating direction method of the multipliers (ADMM) \citep{admm_boyd}, which is a framework for decomposing optimization problems into a sequence of parallel tasks. Using ADMM, we show that the task of debiasing multiclass datasets reduces to a sequence of parallel debiasing jobs on each class separately, along with an aggregation step. Each debiasing job can be executed using the randomized threshold optimizer (RTO) algorithm of \citet{alabdulmohsin2021}, which was originally proposed as a post-hoc rule for binary classification -- in this work, we adopt it as a preprocessing method. The overall algorithm inherits the guarantees of ADMM including convergence and optimality. Our empirical results demonstrate that the proposed algorithm can lead to an improvement over the two outlined baselines; i.e. treating multiclass problems as multi-label, and transforming the features instead of the labels. 

Surprisingly, we also demonstrate that the baseline multi-label approach yields competitive results in most (but not all) settings despite the potential impact of normalization on bias (cf. Figure \ref{fig:multilabel_fails}). In Appendix \ref{appendix:ml}, we provide an argument for why this can happen under idealized assumptions.


\paragraph{Statement of Contribution.}This work addresses a gap in the machine learning literature on demographic parity for multiclass classification with non-binary sensitive attributes. 
Our contributions are: (1) we derive a method for debiasing multiclass datasets with categorical sensitive attributes of arbitrary cardinality with respect to demographic parity -- our approach reduces to a sequence of debiasing tasks on binary labels, (2) we establish theoretical guarantees for the proposed algorithm, (3) we study the impact of the experiment settings (e.g. number of features, number of classes, etc) on the debiasing algorithms using synthetic data and validate different debiasing methods on real-world datasets from three domains: social science, computer vision, and healthcare, and (4) we evaluate the baseline multi-label approach in debiasing multiclass datasets and demonstrate that it yields competitive results in most (but not all) settings.

\section{Related Work}\label{sect::relatedworks}
In the \emph{binary} classification setting, several algorithms have been proposed for mitigating bias in machine learning. These are often classified into three groups depending on which step in the machine learning pipeline the debiasing effort takes place. First, there are \emph{preprocessing} methods that are applied prior to training, such as by learning a fair representation \citep{pmlr-v28-zemel13,lum2016statistical,bolukbasi2016man,calmon2017optimized,pmlr-v80-madras18a} or re-weighting training examples \citep{kamiran2012data}. One example of a preprocessing method that transforms the labels is the optimized score transformation (OST) method \citep{wei2019optimized} whereas the DP remover method of \citet{feldman2015certifying} transforms the features. 
Feature debiasing has the advantage of not depending on the labels and can therefore be applied in any task setting. 

Second, \emph{in-processing} methods intervene during training, such as by adjusting the gradient updates \citep{zhang2018mitigating} or by adding explicit constraints into the optimization problem; e.g. \citep{JMLR:v20:18-262}.  In \citet{pmlr-v80-agarwal18a}, it is shown that many fairness criteria can be enforced during training via a reduction approach to cost-sensitive classification rules. Our proposed algorithm for the multiclass setting also reduces to a sequence of debiasing rounds. However, every round decomposes into multiple parallel debiasing jobs for each label separately (i.e. using algorithms for debiasing binary labels) and we operate in the preprocessing setting. Reduction methods, in which solutions to simple problems are reused to solve complex tasks, are not uncommon in machine learning. Besides \citet{pmlr-v80-agarwal18a}, other examples include the \emph{MetaCost} method \citep{domingos1999metacost}, error correcting codes \citep{dietterich1994solving},  boosting \citep{boostingbook}, conditional probability estimation \citep{beygelzimer2014conditional}, ranking \citep{balcan2007robust}, and relating reinforcement learning to classification \citep{langford2005relating}.

Third, many post-processing methods have been proposed in the literature \citep{corbett2017algorithmic,menon2018cost,celis2019classification,kamiran2012decision,hardt2016equality,wei2019optimized,alabdulmohsin2021}. One recent example is the randomized threshold optimizer (RTO) of \citet{alabdulmohsin2021}, which can provably approximate the optimal unbiased predictor (i.e. it is statistically consistent) and can be solved efficiently via stochastic gradient descent (SGD). Our ADMM based approach for debiasing multi-class datasets utilizes RTO to debias every label separately before aggregating results. 

To our knowledge, two methods have been developed for multiclass classifications. First is the algorithm of \citet{denis2021fairness}, which assumes that the sensitive attribute is binary, whereas the sensitive attribute can be non-binary in our setting. Second, \citet{yang2020fairness} propose both in-processing and post-processing procedures; the latter of which requires access to the sensitive attribute at prediction time and cannot be directly applied as a preprocessing method on discrete labels.

\paragraph{Advantages of R2B.}The proposed ADMM-based reduce-to-binary (R2B) algorithm is a pre-processing method. This offers three immediate advantages. First, it is agnostic to the choice of the training algorithm; unlike, for example, in-processing methods that are often designed with a specific model and a choice of training method in mind. Second, R2B does not require access to the sensitive attribute at prediction time, which is a critical advantage over  post-processing methods \citep{JMLR:v20:18-262}. Third, R2B reduces the debiasing task to a sequence of  rounds of \emph{debiasing} the labels, not training classifiers. The computational overhead in our approach is often negligible compared to in-processing methods that provide a reduction approach to a sequence of classification rules, such as in \citet{pmlr-v80-agarwal18a}, in which the entire model is \emph{trained} several times. Unlike the multi-label approach, R2B is guaranteed to debias the dataset up to the prescribed bias tolerance level. In addition, R2B performs better than other baselines, such as preprocessing the features instead of the labels. Importantly, it can be applied in the multiclass setting with a non-binary sensitive attribute.

\section{Reduction to Binary Method}
\paragraph{Notation.}We reserve boldface letters for random variables (e.g. $\textbf{x}$), small letters for instances (e.g. $x$), capital letters for matrices (e.g. $X$), and calligraphic typeface for sets (e.g. the instance space $\mathcal{X}$). If $f:\mathcal{X}\to\mathbb{R}^n$ is a multivariate function, we write $f_k(x):\mathcal{X}\to\mathbb{R}$ for the $k$-th component of $f$. We denote $[n] = \{0, 1,\ldots,n-1\}$ and reserve $\eta_k(x)$ for the Bayes regressor: $\eta_k(x) = p(\textbf{y}=k|\textbf{x}=x)$. We assume that the sensitive attribute is known at training time and it has a finite range. We denote the instance space $\mathcal{X}$, the dataset $\mathcal{D}: |\mathcal{D}|=N$, the target set $\mathcal{Y}=\{0, 1,\ldots,L-1\}$ and the sensitive attribute $g:\mathcal{X}\to\mathcal{S}$ where $\mathcal{S}=[R]$. We write $\mathcal{X}_\mathcal{S}$ for the portioning of $\mathcal{X}$ induced by $\mathcal{S}$; i.e. $\mathcal{X}_\mathcal{S}=\{\mathcal{X}_0,\ldots,\mathcal{X}_{R-1}\}$ is the set of groups/subpopulations where $\mathcal{X}=\cup_{s=0}^{R-1}\mathcal{X}_s$. We will occasionally write $x_i$ for the $i$-th instance in the training dataset. We also denote $\one\in\mathbb{R}^n$ for the vector of all 1's and $\zero\in\mathbb{R}^m$ for the vector of all zeros. In both cases, the dimension is implicit and should be inferred with ease from the context. 
Finally, $||X||_F$ is the Frobenius norm of the matrix $X$.

\subsection{Multiclass Demographic Parity}
\paragraph{Definition.}Before deriving the reduction-to-binary (R2B) method, we first describe how demographic parity is extended to the multiclass setting. In the binary classification setting, demographic parity measures the difference in \emph{mean} outcomes across groups. More precisely, let $f:\mathcal{X}\to[0, 1]$ be a binary predictor that outputs a probability score $f(x)=p(\by=1|\bx=x)$. Then, $f$ is said to satisfy $\epsilon$ demographic parity if the following holds \citep{dwork2012fairness,zafar2017fairness,mehrabi2019survey,alabdulmohsin2021}:
\begin{equation}
 \max_{s\in\mathcal{S}}\,\mathbb{E}_{\bx}[f(\bx)\,|\,g(\bx)=s] \,-\,\min_{s\in\mathcal{S}}\,\mathbb{E}_{\bx}[f(\bx)\,|\,g(\bx)=s] \le \epsilon,
\end{equation}
where $g:\mathcal{X}\to\mathcal{S}$ is the sensitive attribute. A crowd-sourcing study found that DP matches with the common perception of bias \citep{srivastava2019mathematical}. In the multiclass setting, let $f_k:\mathcal{X}\to[0,1]$ be the probability score assigned to the class $k\in\mathcal{Y}$ given the instance $x\in\mathcal{X}$ and let $f:\mathcal{X}\to[0,1]^L$ be the multivariate function $f(x) = (f_0(x),\,f_1(x),\ldots,f_{L-1}(x))^T$. In this work, we follow the definition used in \citet{denis2021fairness} and say that $f$ satisfies $\epsilon$ demographic parity if:
\begin{equation}\label{eq:multiclass:dp}
    DP(f) \doteq \max_{k\in\mathcal{Y}}\;\big\{ \max_{s\in\mathcal{S}}\,\mathbb{E}_{\bx}[f_k(\bx)\,|\,g(\bx)=s] \,-\,\min_{s\in\mathcal{S}}\,\mathbb{E}_{\bx}[f_k(\bx)\,|\,g(\bx)=s]\big\} \le \epsilon.
\end{equation}

Hence, predictors with small demographic parity have similar mean outcomes across all groups $\mathcal{S}$ with respect to \emph{all} of the predicted targets. We take the maximum here instead of the average to avoid pitfalls that can arise when some classes are more preferred than others, which is analogous to the \lq\lq fairness gerrymandering'' phenomenon, where predictors may exhibit small bias across the intersection of groups \emph{on average}, but not at the worst-case intersection ~\citep{kearns2018preventing}. Here, instead of disparities across groups, we highlight worst-case disparities across {\em classes}. In Appendix \ref{appendix:scaling}, we discuss some scaling effects of the number of classes on the definition of multiclass DP.

\subsection{Derivation of the Reduction-to-Binary (R2B) Algorithm}\label{sect::derivation}
In the binary classification setting where $\mathcal{Y}=\{0,\,1\}$, a natural measure of performance is the 0-1 misclassification error rate. Minimizing the expected 0-1 error is equivalent to maximizing the linear functional $\mathbb{E}[\hat y(\bx)(2\eta_1(\bx)-1)]$ \citep{alabdulmohsin2021}, where $\hat y$ are the new debiased labels and $\eta_1(x) = p(\by=1|\bx=x)$ is the Bayes regressor. Here, the minimization is with respect to the debiased labels $\hat y:\mathcal{X}\to\mathbb{R}^L$. Hence, if access to the Bayes regressor is available at training time, one can usually compute  $\hat y(\bs)$ that minimizes the expected 0-1 loss subject to the desired fairness constraints by solving a convex optimization problem; cf. \citep{menon2018cost,alabdulmohsin2021,celis2019classification,wei2019optimized}. 

Unfortunately, such an advantage no longer holds in the multiclass setting because the top-1 accuracy is often insufficient. For top-k accuracy, however, the squared loss is statistically consistent  \citep{pmlr-v119-yang20f}. Hence,  we propose to minimize the $\ell_2$ distance $\mathbb{E}[(\hat y(\bx)-y(\bx))^2]$ between the debiased predictions and the original labels. Throughout the sequel, $y(x):\mathcal{X}\to\mathbb{R}^L$ gives the probability scores assigned to the different classes for the instance $x\in\mathcal{X}$, which can be a degenerate distribution as is often the case. The objective function $\mathbb{E}[(\hat y(\bx)-y(\bx)^2)]$ can be written as:
\begin{equation}\label{eq::objective}
    (\lambda/2)\,\mathbb{E}_{\bx}\,||\hat y(\bx)||_2^2 \;-\; \mathbb{E}_{\bx}[\hat y(\bx)^Ty(\bx)] \;+\; \mathrm{Constant},
\end{equation}
where $\lambda=1$. Rewriting the objective function in the form (\ref{eq::objective}) provides an alternative interpretation: by minimizing the objective in (\ref{eq::objective}), one seeks a solution that minimizes the \emph{regularized} top-1 error. This holds because the probability of correctly classifying an example $x$ equals $\hat y(x)^Ty(x)$ when $y_k(x) = p(\by=k|\bx=x)$. Regularization, however, discourages extreme predictions. We fix $\lambda=1$ since it corresponds to minimizing the squared loss, although R2B can handle any $\lambda>0$.

The task of debiasing labels in a dataset $\mathcal{D}$ can hence be cast into the convex optimization problem:
\begin{align}
    \label{eq:old_obj}&\min_{\hat y:\mathcal{X}\to\mathbb{R}^L}\;&&\sum_{x\in\mathcal{D}}\Big\{(1/2)\, ||\hat y(x)||_2^2 \;-\; \hat y(x)^Ty(x)\Big\}.\\
    \nonumber&\text{s.t.} &&    \max_{k\in\mathcal{Y}}\;\big\{ \max_{s\in\mathcal{S}}\,\mathbb{E}_{\bx}[\hat y_k(\bx)\,|\,g(\bx)=s] \,-\,\min_{s\in\mathcal{S}}\,\mathbb{E}_{\bx}[\hat y_k(\bx)\,|\,g(\bx)=s]\big\} \le \epsilon.
\\
    \nonumber& &&\forall x\in\mathcal{D}:\;\one^T\hat y(x) = 1 \quad \wedge \quad \hat y(x) \ge \zero.
\end{align}
Instead of solving this optimization problem directly, we re-express it into a more convenient form:
\begin{align}
    \label{eq:new_obj}&\min_{\hat y:\mathcal{X}\to\mathbb{R}^L}\;&& \sum_{x\in\mathcal{D}}\,\Big\{(1/2)\,||\hat y(x)||_2^2 \;-\; \hat y(x)^Ty(x) \;+\;  \mathbb{I}_{\mathcal{F}}(z(x))\Big\}.\\
    \nonumber&\text{s.t.} &&    \max_{k\in\mathcal{Y}}\;\big\{ \max_{s\in\mathcal{S}}\,\mathbb{E}_{\bx}[\hat y_k(\bx)\,|\,g(\bx)=s] \,-\,\min_{s\in\mathcal{S}}\,\mathbb{E}_{\bx}[\hat y_k(\bx)\,|\,g(\bx)=s]\big\} \le \epsilon.
\\
    \nonumber& &&\forall x\in\mathcal{D}:\;\hat y(x) = z(x)\quad \wedge \quad \zero \le \hat y(x) \le \one.
\end{align}
We introduced a dummy constraint $\hat y(x)\le \one$ and replaced  $\one^T\hat y(x)=1$ with a term $\mathbb{I}_{\mathcal{F}}(\hat y(x))$ in the objective function where:
\begin{equation}
    \mathbb{I}_{\mathcal{F}}(\hat y(x)) = \begin{cases}
    0,      & \one^T\hat y(x)=1\\
    \infty, & \text{otherwise}.
    \end{cases}
\end{equation}
Clearly, the optimization problem (\ref{eq:new_obj}) is equivalent to the original optimization problem (\ref{eq:old_obj}).

Algorithm \ref{algorithm} presents pseudocode for the proposed reduction-to-binary (R2B) method for solving the optimization problem in (\ref{eq:new_obj}). At a high-level, the algorithm decomposes the debiasing task into a sequence of rounds. In each round, the probability scores assigned to different classes are debiased \emph{independently}. This produces a new set of scores, denoted $H^{(t)}$, that are not normalized. The next step is to normalize them so that they sum to one.  The set of normalized probability scores is given by $Z^{(t)}$. After that, the objective function is altered slightly and the process is repeated. We prove that Algorithm \ref{algorithm} returns the \emph{optimal} solution to (\ref{eq:new_obj}) and discuss suitable stopping criteria in Section~\ref{sect::analysis}.


\begin{algorithm}[t]
 \small
 \vspace{1mm}
 \textbf{Inputs:} (1) Step size $\tau>0$; (2) Demographic parity tolerance level $\epsilon\ge 0$; (3) Label matrix $Y\in\mathbb{R}^{N\times L}$, where $Y_{ik}=p(\by=k)$ for the $i$-th training example; (4) Sensitive attribute $g\in[R]^N$, where $g(i)$ is the sensitive class of the $i$-th training example.
 
 \vspace{1mm}
 
 \textbf{Output:}
 Debiased probability scores $\hat Y\in\mathbb{R}^{N\times L}:\,\hat Y\ge \zero 
 \wedge \hat Y\one = \one$ that minimize $||Y-\hat Y||_F$ subject to $\epsilon$ demographic parity; i.e. a solution to  Equation (\ref{eq:new_obj}).
 
 \vspace{1mm}
 \textbf{Training:} Set $Y^{(0)} = Y$ and $Z^{(0)}=U^{(0)}=\zero\in\mathbb{R}^{N\times L}$. Repeat until stopping criterion (cf. Section \ref{sect::analysis}):\vspace{1mm}
 \begin{enumerate}
     \item \emph{Debias in Parallel:}\\Set $F^{(t)} = Y + \tau\,(Z^{(t)}-U^{(t)})$. Let $f_k^{(t)}\in\mathbb{R}^N$ be the $k$-th column of $F^{(t)}$.  Solve the following  debiasing task for each class separately (e.g. using RTO \citep{alabdulmohsin2021}):
     \begin{align*}
         &\hat y_k^{(t)} = &&\arg\min_{\zero\le \hat y_k\le \one}\; &&(1+\tau)/2||\hat y_k||_2^2 - \hat y_k^T\, f_k^{(t)}\\
         &
         &&\text{s.t.} &&\max_{s\in\mathcal{S}}\,\mathbb{E}_{\bi\in[N]}[\hat y_k(\bi)\,|\,g(\bi)=s] \,-\,\min_{s\in\mathcal{S}}\,\mathbb{E}_{\bi\in[N]}[\hat y_k(\bi)\,|\,g(\bi)=s] \le \epsilon
     \end{align*}
    \item \emph{Aggregate Results:}
    \begin{enumerate}
        \item Set $H^{(t+1)}_{ki} = \hat y_k^{(t)}(i)$: unnormalized score assigned to class $k\in\mathcal{Y}$ for the $i$-th training example.
        \item Normalize probability scores using:
        \begin{equation*}
            Z^{(t+1)} = H^{(t+1)} + U^{(t)} - \frac{1}{L}\,\Big[(H^{(t+1)}+U^{(t)})\cdot \textbf{1} - \textbf{1}\Big]\cdot \textbf{1}^T.
        \end{equation*}
        \item Update: $U^{(t+1)} = U^{(t)} + H^{(t+1)} - Z^{(t+1)}$.
    
    \end{enumerate}
 \end{enumerate}\vspace{1mm}
 \noindent \textbf{Return} $\hat Y = Z$. \vspace{3mm}
 \caption{Pseudocode of the reduction-to-binary (R2B) algorithm for debiasing multiclass datasets with categorical sensitive attributes of arbitrary cardinality.}\label{algorithm}
\end{algorithm}

\subsection{Analysis of the Algorithm}\label{sect::analysis}
\begin{proposition}[Optimality]\label{prop:optimality}
Algorithm \ref{algorithm} terminates with an optimal solution to  (\ref{eq:new_obj}), in which the debiased probability scores assigned to the $i$-th example are given by the $i$-th row of the matrix $\hat Y$.
\end{proposition}

\begin{figure}
    \centering
    \includegraphics[width=\columnwidth]{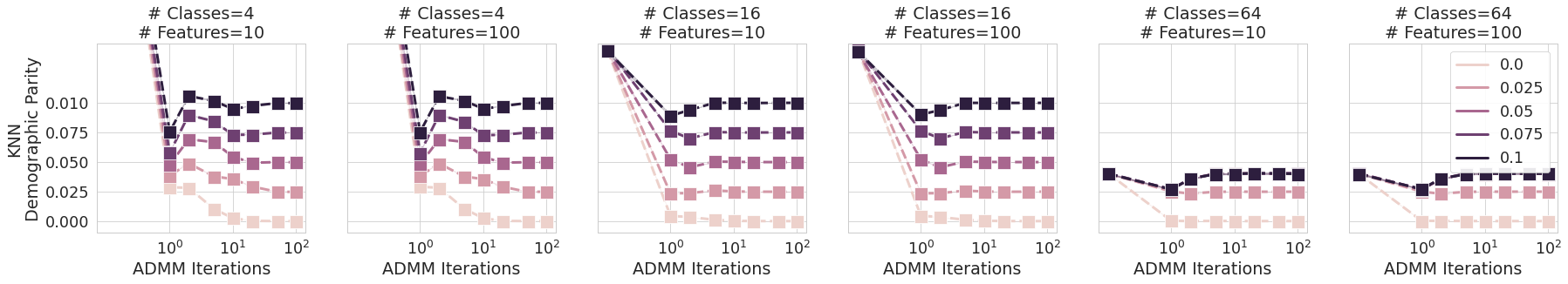}
    \caption{The level of bias in the \emph{training} data is plotted  as measured by (\ref{eq:multiclass:dp}) vs. the number of ADMM rounds for different levels of bias tolerance $\epsilon\in\{0, 0.025, 0.05, 0.075, 0.10\}$ (see legend) using the $k$NN classifier (see full figures in Appendix \ref{appendix:synth_full}). In general, 50-100 rounds of ADMM in R2B are sufficient to reach convergence in agreement with prior observations \citep{admm_boyd}.}
    \label{fig:synth_addconvergence}
\end{figure}

\paragraph{Stopping Criteria.}The simplest stopping criterion to use in Algorithm \ref{algorithm} is the number of ADMM rounds. \citet{admm_boyd} observes that 50-100 rounds are often sufficient, which we also observe to be true in our experiments (see Figure \ref{fig:synth_addconvergence}). Alternatively, one may use the sum of the \lq\lq primal" and \lq\lq dual residuals". In Algorithm \ref{algorithm}, the optimality conditions of ADMM \citep{admm_boyd} reduce to the following conditions $Z^{(t)} = H^{(t)}$ and $Z^{(t+1)} = Z^{(t)}$. Hence, we may define $\delta_p = ||Z^{(t)} - H^{(t)}||_F$ and $\delta_r = ||Z^{(t+1)} - Z^{(t)}||_F$ and stop when $\delta_p+\delta_r$ are below a prescribed threshold. In our implementation, we follow the first approach and set the maximum number of ADMM rounds to 100.

\paragraph{Bias Guarantee.}For our next result, we write $\hat{\mathbb{E}}[f(\bx)]$ for the average value of a function $f$ on the training examples, and write $\mathbb{E}[f(\bx)]$ for the expectation of $f$ over the true distribution of instances. 

\begin{proposition}[Bound on bias]\label{prop:bias}
Let $\mathbb{P}^L$ be the probability simplex in $\mathbb{R}^L$ and $\mathcal{F}$ be a class of functions from $\mathcal{X}$ to $\mathbb{P}^L$, such that the set $\{f_k: f\in\mathcal{F}\}$ has a Rademacher complexity $d$. Suppose that all training examples $(\bx,\by,\bs)\in\mathcal{D}$ are drawn i.i.d. and are debiased using Algorithm \ref{algorithm} with bias level $\epsilon\ge 0$. Let $h(\bs):\mathcal{X}\to\mathbb{P}^L$ be the debiased labels. Let $f\in\mathcal{F}$ be the final classifier. Then, with a probability of at least $1-\delta$, we have: $DP(f) \le \epsilon \,+\, 2\tau \,+\, 2d \,+ \,\sqrt{\frac{\log\frac{2LR}{\delta}}{2n_0}}$,
where $n_0 = \min_{s\in[R]}\,\big|\big\{\mathbb{I}\{g(\bx)=s\}\,:\,\bx\in\mathcal{D}\big\}\big|$ is the size of the smallest group in the training dataset, $L=|\mathcal{Y}|$ is the number of classes, $R=|\mathcal{S}|$ is the number of groups, and $\tau = \sup_{k\in\mathcal{Y},\,s\in[R]} \big|\hat{\mathbb{E}}[f_k(\bx)\,|\,g(\bx)=s] - \hat{\mathbb{E}}[h_k(\bx)\,|\,g(\bx)=s]\big|$,
where both expectations are measured on the training dataset.
\end{proposition}

Proposition \ref{prop:bias} provides a formal justification to the preprocessing approach; it states that if the training labels are debiased using Algorithm \ref{algorithm}, then a classifier trained on the debiased data is guaranteed to exhibit small bias in the future as long as four conditions are satisfied: (1) the level of bias in the training data is small; i.e. $\epsilon\ll 1$ in Algorithm \ref{algorithm}, (2) the classifier fits the training examples well; i.e. $\tau\ll 1$, (3) the complexity of the classifier is not too large (does not memorize examples); i.e. $d\ll 1$, and (4) there exists a large number of training examples for each group in $\mathcal{S}$; i.e. $n_0\gg \log LR$.

\section{Experiments}\label{sect::exp}
We compare R2B against the two baselines: (1) treating multiclass datasets as multi-label, and (2) transforming the features. In the multi-label approach, we use two recent algorithms for debiasing binary labels: optimized score transformation (OST) \citep{wei2019optimized} and  RTO  \citep{alabdulmohsin2021} with $\gamma=0.01$ and $\rho=\mathbb{E}[\by_k]$ (default) values. After that, the scores are normalized to sum to one. Note that $\gamma\ll 1$ in the RTO algorithm corresponds approximately to a \emph{hard}-thresholding rule, hence we refer to it as hard-threshold optimizer (HTO). In the second approach, we apply the demographic parity remover (DPR) method of \citet{feldman2015certifying},  which debiases features. In the latter case, we use equal-mass binning when estimating the cumulative density function (see Appendix \ref{appendix:dpremover}). All three baselines can be motivated via provable optimality guarantees \citep{wei2019optimized,alabdulmohsin2021,feldman2015certifying}. We also include for comparison training without debiasing the data, which we denote as BL\footnote{Source code is publicly available at: \url{https://github.com/google-research/google-research/tree/master/ml_debiaser}}. We note that existing multiclass fairness approaches such as \citet{yang2020fairness} cannot be applied for pre-processing as they require either probability estimates or weighted classifiers. We highlight in \textbf{boldface} the method with the best accuracy among the set of methods that achieve the smallest bias within the margin of error.

In R2B, on the other hand, we use a step size of $\tau=0.5$ in Algorithm \ref{algorithm} and a maximum number of 100 ADMM rounds. We also report results when a \emph{single} round of R2B is used, which we denoted as R2B\textsubscript{0}. 
Except for the healthcare dataset whose splits are fixed, we split data at random into 25\% for test and 75\% for training. We debias the dataset prior to training and measure performance (e.g.  accuracy and DP) on the test split. All methods use the same splits. We re-run every experiment with ten different random seeds and report both the averages and 99\% confidence intervals.

\begin{figure}
    \centering\footnotesize
    \includegraphics[width=\columnwidth]{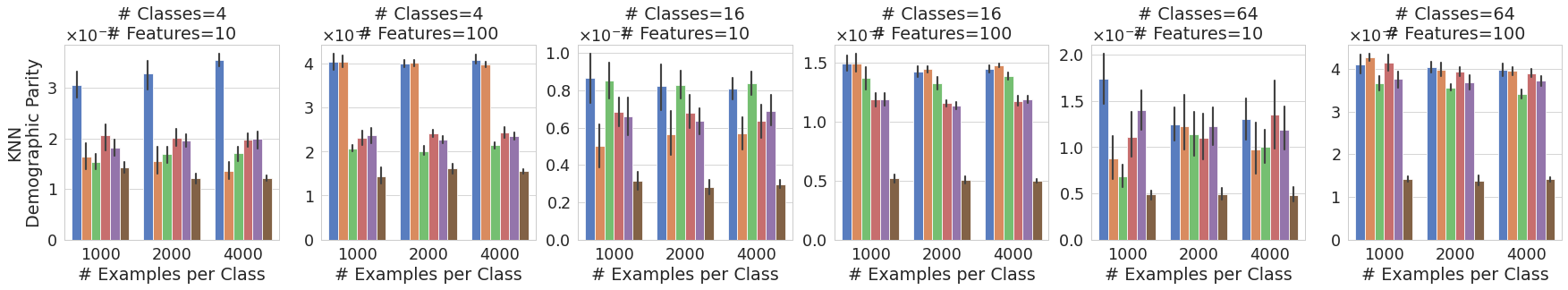}\\[2mm]
    \includegraphics[width=\columnwidth]{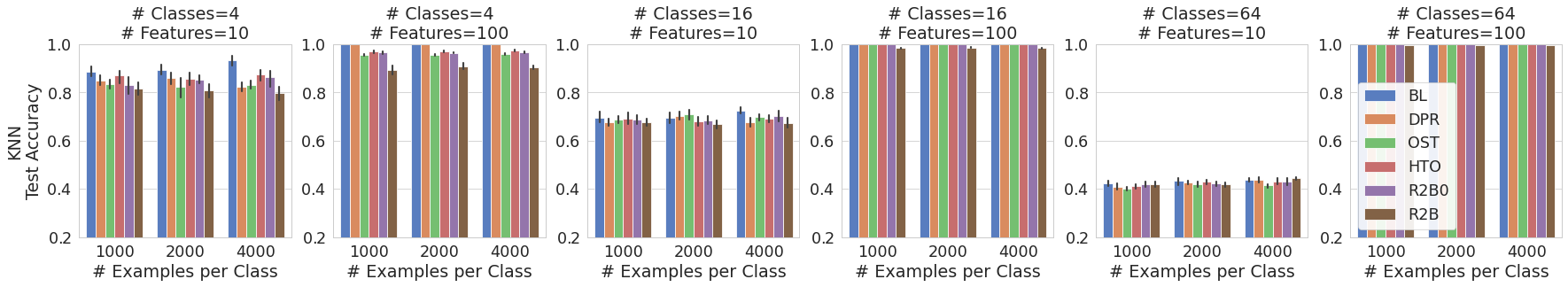}
    \caption{Top two displays the multiclass demographic parity (DP) as measured using (\ref{eq:multiclass:dp}) for the synthetic dataset with varying numbers of classes, features, and training examples using $k$NN as a classifier (see Appendix \ref{appendix:synth_full} for full figures). The reduction-to-binary (R2B) method provides a stronger fairness guarantee than the competing methods. The bottom row shows the prediction accuracy.}
    \label{fig:synth_bias_and_acc}
\end{figure}

\subsection{Synthetic Dataset}
We begin our analysis with synthetic data. In this dataset, let $L$ be the number of classes and $d$ be the number of features. We use a mixture of $L$ Gaussians whose means are sampled from $\mathcal{N}(0, \sigma^2I)$ with $\sigma^2=1/d$ and $\mathcal{Y}=[L]$. For the sensitive attribute, we set it equal to $\mathbb{I}\{\by=0\}$ with probability $1/2$; otherwise, it is chosen uniformly at random from the set $\{0,\ldots, 4\}$; i.e. $|\mathcal{S}|=5$. Hence,  bias is introduced into the dataset. The classifiers are $k$-NN  with $k=5$ and Random Forest (RF) with maximum depth 5, both implemented using Scikit-Learn \citep{scikit-learn}. We vary the number of classes in $\{4,\, 16,\, 64\}$, the number of features in $\{10,\, 100\}$, and the number of training examples per class in $\{1000,\, 2000,\, 4000\}$. We set the bias tolerance $\epsilon$ to zero in all experiments since, otherwise, these methods provide incomparable mechanisms for controlling bias. Figure \ref{fig:synth_bias_and_acc} displays the results for $k$-NN and Appendix \ref{appendix:synth_full} contains the full figures.

As shown in Figure \ref{fig:synth_bias_and_acc}, R2B provides a stronger fairness guarantee than the other competing methods. In addition, we observe that debiasing the features performs poorly when the number of features is large. As mentioned earlier, one possible explanation is that as the number of features grows, the rate of convergence of the multivariate empirical cumulative distribution function (EDF) to the true cumulative distribution function (CDF) decreases \citep{naaman2021tight}. However, debiasing the features in \citet{feldman2015certifying} relies on having a correct estimate of the CDF since it matches  quantiles. 

Moreover, we observe that the multi-label approach (i.e. OST, HTO, and R2B0) can also fail in several cases. As illustrated in the cartoon example of Figure \ref{fig:multilabel_fails}, this is because normalization impacts bias. In one experiment, for example, the level of DP in the training labels is reduced  to less than 0.01 after debiasing labels separately using OST but bias increases to over 0.22 following normalization.   

In Figure \ref{fig:synth_addconvergence}, we also plot the level of DP in the \emph{training} labels as a function of the number of ADMM rounds in R2B for five different levels of bias tolerance $\epsilon\in\{0,\, 0.025,\, 0.05,\, 0.075,\, 0.10\}$. We observe that R2B converges in about 50 ADMM iterations, which is consistent with earlier observations in the literature \citep{admm_boyd}. In addition, when the level of bias in the data is smaller than $\epsilon$, the algorithm does not introduce any additional bias as expected.

\begin{table*}
    \centering
    \scriptsize
    \caption{A summary of the performance results (99\% confidence intervals) for the five debiasing algorithms on the Adult Income benchmark dataset, where the goal is to predict the education level. For each of the two classifiers Random Forest (RF) and $k$-NN, we either use the original data ($p=0$) or inject bias ($p=0.5$) as described in Section \ref{sect:exp_real}. First column is the bias in the data. The debiasing methods are the baseline (BL), DP remover (DPR) \citep{feldman2015certifying}, optimized score transformation (OST) \citep{wei2019optimized}, randomized threshold optimizer with $\gamma=0.01$ (HTO) \citep{alabdulmohsin2021}, R2B\textsubscript{0}, and R2B (our method).
    }\vspace{1mm}
    
    \begin{tabular}{lll|ccccccc}
    \toprule
      Criterion& $p$ &Learner &DATA & BL & DPR & OST  & HTO &
         R2B\textsubscript{0} &
         R2B (ours) \\
         \midrule
           \multirow{4}{*}{DP [\%]} &
0.0&\sc rf & $5.0$&${3.8\pm.2}$&$\bm{1.5\pm.3}$&${3.7\pm.3}$&${2.9\pm.3}$&${2.8\pm.3}$&${2.2\pm.2}$\\
&&\sc knn &$5.0$ &${4.7\pm.5}$&${3.0\pm.7}$&${4.5\pm.5}$&${3.2\pm.5}$&${3.1\pm.5}$&$\bm{2.1\pm.4}$\\
&0.5&\sc rf &$48.0$ &${22.8\pm.3}$&${9.5\pm.4}$&${1.7\pm.3}$&${2.2\pm.3}$&${2.1\pm.2}$&$\bm{1.0\pm.2}$\\
&&\sc knn &$48.0$ &${26.5\pm.5}$&${23.3\pm.6}$&${2.9\pm.4}$&${3.4\pm.6}$&${3.3\pm.7}$&$\bm{1.1\pm.2}$\\\midrule

  \multirow{4}{*}{ACC [\%]} &
0.0&\sc rf &$\star$ &$42.6\pm.6$&${41.0\pm.7}$&$42.6\pm.6$&$42.5\pm.6$&$42.6\pm.5$&$42.5\pm.6$\\
&&\sc knn &$\star$&$35.5\pm.5$&$33.4\pm.3$&$35.3\pm.5$&$35.2\pm.5$&$35.2\pm.5$&${35.5\pm.5}$\\
&0.5&\sc rf &$\star$&$42.0\pm.7$&$39.0\pm.7$&$34.1\pm.4$&$33.9\pm.6$&$34.0\pm.5$&${34.0\pm.4}$\\
&&\sc knn &$\star$&$34.9\pm.5$&$34.3\pm.6$&$30.2\pm.4$&$29.4\pm.4$&$29.6\pm.7$&${31.1\pm.4}$\\\bottomrule
   \end{tabular}
    \label{tab:postprocessing_adult_default}
    \vspace{-2mm}
\end{table*}

\subsection{Real-world Applications}\label{sect:exp_real}
Next, we validate R2B on applications from three domains: (1) social science, (2) computer vision, and (3) healthcare. Experiments involving neural networks are executed on Tensor Processing Units.

\paragraph{Adult Income dataset.}
The first dataset is the Adult Income dataset \citep{kohavi1996scaling}. In our case, we predict the education level of each individual from the remaining attributes. There are 16 classes in this dataset and 29 features, such as marital status, age, and occupation. The total number of examples (both training and test) is 48,842. The sensitive attribute is sex and only includes two categories (`Male', `Female'). We compare the debiasing methods in Table \ref{tab:postprocessing_adult_default} for both the Random Forest (RF) and $k$-NN classifiers (top two rows). Besides the original dataset, we also introduce bias and class imbalance into the data (both training and test) and compare methods: with probability $p=0.5$, the label $\by\in\{0,1,\ldots,15\}$ is set to be equal to the sensitive attribute $\bs\in\{0,1\}$. This increases bias in the original labels from about 0.05 to around 0.48 and introduces class imbalance as well\footnote{The level of demographic parity in the original data can be different from the baseline results, because the baseline model may not predict the original labels. For instance, when $p=0.5$, noise is added to the labels to make them biased but such noise can be ignored by the model.}. The performance of each debiasing method in this dataset is shown in Table \ref{tab:postprocessing_adult_default}, bottom two rows. In most cases, R2B provides a stronger bias guarantee compared to the other  methods. Appendix \ref{app_topk} provides similar performance results for Top-K accuracy.



\paragraph{COCO dataset.}
The second dataset we use is the COCO dataset \citep{DBLP:journals/corr/LinMBHPRDZ14}. It contains 80 classes corresponding to the objects in each image, such as chairs, cars, and handbags. We transform this multi-label  problem into a multiclass problem using \emph{soft} labels: we set the target label $\by$ to be equal to the \emph{fraction} of the objects that belong to each class in each image. The soft label approach corresponds formally to  the task of predicting the \emph{distribution} of objects seen in the image or, equivalently, the task of predicting the class of an object chosen uniformly at random from the corresponding image.
We follow the procedure of \citep{wang2020revise} in inferring the sensitive attribute based on the image caption: we use images that contain either the word \lq\lq woman" or the word \lq\lq man" in their captions but not both. The total number of examples is 22,616. Because the task here is to predict the distribution of classes seen in the image, we define accuracy in terms of the total variation distance between probability distributions. Specifically, let $\hat\by$ be the model's prediction, then $\mathrm{acc}(\hat\by, \by) = 1 - (1/2) ||\hat\by-\by||_1$. Note that $0\le \mathrm{acc}(\hat\by,\by)\le 1$. See Appendix \ref{appendix:coco_training} for further details about the training procedure. Besides the original dataset, we also introduce bias and class imbalance with $p=0.5$ as described previously. Results are  provided in Table \ref{tab:postprocessing_coco}. As shown in the table, both R2B and the multi-label methods perform much better than transforming the features (DPR), which is consistent with the earlier observations on synthetic data. Moreover, R2B performs better overall. 

\begin{table*}
    \centering
    \scriptsize
    \caption{
A summary of the performance results (99\% confidence intervals) for the five debiasing algorithms on the COCO benchmark dataset with  soft labels. See Section \ref{sect:exp_real} for details. Similar to Table \ref{tab:postprocessing_adult_default}, we experiment with both the original dataset ($p=0$) and with introduced bias ($p=0.5$).
    }\vspace{1mm}
    
    \begin{tabular}{ll|ccccccc}
    \toprule
        & $p$ & DATA & BL & DPR & OST  & HTO &
         R2B\textsubscript{0} &
         R2B (ours) \\
         \midrule
           \multirow{2}{*}{DP [\%]} &
0.0& $6.2$&${2.6\pm.4}$&${6.4\pm1.8}$&${2.2\pm.3}$&${3.8\pm.7}$&${2.2\pm.4}$&$\bm{2.4\pm.4}$\\
&0.5& $52.9$
&${4.0\pm.6}$
&${52.3\pm3.4}$
&${0.7\pm.2}$
&${1.3\pm.2}$
&${1.0\pm.4}$
&$\bm{1.1\pm.4}$\\
\midrule 
           \multirow{2}{*}{ACC [\%]} &
0.0& $\star$& $40.4\pm.2$ & $28.0\pm2.0$ & ${39.8\pm.2}$ & $32.8\pm.1$ & ${40.6\pm.2}$ & $\bm{41.3\pm.2}$\\
&0.5& $\star$ 
&$43.8\pm.3$
&$50.1\pm2.7$
&${42.5\pm.2}$
&$39.6\pm.2$
&${43.8\pm.2}$
&$\bm{48.1\pm.4}$
\\\bottomrule
   \end{tabular}
    \label{tab:postprocessing_coco}
    \vspace{-2mm}
\end{table*}
\paragraph{Dermatology.}
In this task, we are interested in predicting 27 skin conditions (26 plus an `other' label representing the long tail) from images of the pathology of interest, the patient's age and sex\footnote{Sex corresponds to clinician or self recorded sex and only includes two categories (`Male', `Female')}. The dataset is a subset of the one used in \cite{Liu2020-qr} and consists of de-identified retrospective adult cases from a teledermatology service serving 17 sites from 2 states in the United States. It is split according to condition prevalence for training ($n=12,024$), tuning for hyper-parameters ($n=1,925$) and hold-out testing ($n=1,924$). We train a deep learning model to predict skin conditions as a multiclass task, as per the architecture described in \citet{Liu2020-qr,Roy2021-xp}. For debiasing, we fit R2B on the tune split and assess the model performance. Following \citet{Liu2020-qr}, we report the top-1, top-2, and top-3 accuracy on the test split. Table \ref{tab:derm} reports  DP on the test split as well. For ease, we bucket age according to [18,30), [30,45), [45, 65) and [65,90) for debiasing. We note that the number of cases per condition and per intersection of the attributes might be low, as is common in data-scarce domains such as healthcare. Results are presented in Table \ref{tab:derm}.

\newcolumntype{Y}{>{\centering\arraybackslash}X}

\begin{table*}
    \centering
    \scriptsize
    \caption{
    A summary of performance results on a Dermatology dataset \citep{Liu2020-qr}, where the goal is to predict the clinical condition from images of the pathology. 
    DP is 0.09 in the original data. 
    }\vspace{1mm}
    \begin{tabularx}{\columnwidth}{@{}XYYYYYY@{}}
    \toprule
          Metric & BL& DPR & OST & HTO & R2B\textsubscript{0} &R2B  (ours) \\
         \midrule

    DP [\%]&
${11.1\pm1.2}$&${14.3\pm2.2}$&$\phantom{0}\bm{5.1\pm1.0}$&$\phantom{0}{8.0\pm1.4}$&$\phantom{0}\bm{4.8\pm0.5}$&$\phantom{0}\bm{5.0\pm0.5}$\\
    Top-1 [\%]&
${58.6\pm0.5}$&${58.3\pm0.7}$&$\bm{58.5\pm0.5}$&${48.9\pm0.9}$&$\bm{58.3\pm0.7}$&$\bm{59.1\pm0.6}$\\
    Top-2 [\%]&
$\bm{79.1\pm0.7}$&${78.6\pm0.5}$&$\bm{79.3\pm0.6}$&${67.5\pm0.3}$&$\bm{79.0\pm0.1}$&$\bm{79.0\pm0.1}$\\

    Top-3 [\%]&
${89.1\pm0.4}$&${88.8\pm0.6}$&$\bm{88.8\pm0.6}$&${78.4\pm0.8}$&$\bm{87.2\pm0.6}$&$\bm{87.2\pm0.6}$\\


\bottomrule
   \end{tabularx}
    \label{tab:derm}
    \vspace{-2mm}
\end{table*}

\paragraph{Summary of Findings.}
The reduction-to-binary (R2B) approach performs at least as well as the other baselines in all of the experiments and outperforms the other methods in some cases, such as in the synthetic data, the Adult Income dataset, and COCO. In addition, the multi-label approach using OST and R2B\textsubscript{0} seem to offer competitive results in most (but not all) settings. In Appendix \ref{appendix:ml}, we provide an argument for why this might happen under idealized assumptions. However, it is worth emphasizing that the multi-label approach is not guaranteed to debias datasets successfully as illustrated earlier in Figure \ref{fig:multilabel_fails} and demonstrated using  synthetic (Figure \ref{fig:synth_bias_and_acc}) and real (Table \ref{tab:postprocessing_adult_default}) data. On other other hand, R2B offers strong guarantees (cf. Propositions \ref{prop:optimality} and \ref{prop:bias}). 


\begin{table*}
    \centering
    \scriptsize
    \caption{
    A summary of the observed \emph{error parities} for the five debiasing algorithms on all of the classification tasks, which is defined to be the difference between the maximum and minimum losses conditioned on each group. In the original datasets (i.e. $p=0$), we do not observe an increase in error parity when the training data is debiased to account for DP. However, error parity seems to increase when introducing large bias to such data (i.e. with $p=0.5)$. 
    }\vspace{1mm}
    \begin{tabularx}{\columnwidth}{@{}XXYYYYYY@{}}
    \toprule
        p & Task & BL & DPR & OST & HTO & R2B\textsubscript{0} &R2B (ours) \\
         \midrule
         \multirow{5}{*}{0} &
    {Adult {(\sc rf)}} &
$0.7\pm0.1$&$0.7\pm0.4$&$1.0\pm0.2$&$0.8\pm0.2$&$1.0\pm0.2$&$0.7\pm0.3$\\
    & {Adult {(\sc knn)}} 
    &$0.7\pm0.3$&$0.7\pm0.2$&$0.7\pm0.3$&$0.8\pm0.3$&$0.8\pm0.3$&$0.7\pm0.3$\\ 
    & COCO  & $4.4\pm0.5$&$5.7\pm1.0$&$4.0\pm0.5$&$5.1\pm0.6$&$4.2\pm0.5$&$5.1\pm0.5$\\ 
    & {Derm}  & ${5.6\pm1.9}$&${5.1\pm0.8}$&${4.7\pm1.9}$&${5.7\pm2.3}$&${6.1\pm1.8}$&${5.9\pm2.1}$\\ \midrule 
    \multirow{4}{*}{0.5} &
    {Adult {(\sc rf)}} &
$7.6\pm1.1$&$31.5\pm0.8$&$49.9\pm0.2$&$50.1\pm0.3$&$49.9\pm0.4$&$49.8\pm0.3$\\
    & {Adult {(\sc knn)}} &
$2.0\pm0.5$&$1.2\pm0.4$&$12.5\pm0.8$&$8.1\pm0.3$&$7.8\pm0.5$&$31.3\pm0.7$\\ 
    & COCO  & $23.1\pm0.7$&$25.4\pm3.0$&$21.9\pm0.4$&$17.1\pm0.4$&$22.7\pm0.5$&$38.7\pm0.8$\\ 
\bottomrule
   \end{tabularx}
    \label{tab:error_parity}
    \vspace{-2mm}
\end{table*}

\section{Discussion and Limitations}\label{sect:limit}
In this paper, we derive a reduction approach for debiasing multiclass datasets according to demographic parity (DP). The algorithm reduces the overall task into a sequence of parallel debiasing jobs on binary labels. Due to this reduction, the algorithm scales well to large datasets with several classes. We verify empirically on both synthetic and real-world datasets that it outperforms other baselines. 

Nevertheless, it is worth emphasizing that \lq\lq fairness" is a societal concept and should not be reduced to statistical metrics, such as DP \citep{dixon2018measuring,selbst2019fairness}. As such, the claims of this paper hold for the narrow technical definition of DP, not for fairness in its broader sense.

One limitation in R2B is that it accommodates DP but cannot accommodate error parity metrics, such as equalized odds. This is because it operates in the pre-processing setting, where prediction \lq\lq errors" are yet undefined. In particular, for anti-causal predictive tasks \citep{scholkopf2012on} as is the case in dermatology \citep{castro2020causality}, \citet{Veitch2021-rq} suggest that equalized odds would be a causally-grounded fairness constraint.  Nevertheless, in real-world settings, we do not observe an increase in error parity when the training data is debiased to account for DP as shown in Table \ref{tab:error_parity}, except when bias is \emph{introduced} into the data (i.e. with $p=0.5)$). Mitigating DP can reduce accuracy when the labels are correlated with the sensitive attribute as has been noted in several works; e.g. \citep{menon2018cost,zhao2019inherent}. 
Finally, it accommodates \emph{categorical} sensitive attributes only, not continuous. While the majority of protected  attributes are indeed categorical and continuous attributes can be converted into categorical features by binning their values (e.g. age in the dermatology example), handling continuous sensitive attributes without resorting to binning remains a challenge. We leave such questions for future work.





\section*{Acknowledgements}
The authors would like to acknowledge and thank Yuan Liu and Alex Brown at Google Health AI for their support with the dermatology application, and thank Kathy Meier-Hellstern and Lucas Dixon from Google Responsible AI team for their feedback on earlier drafts of this manuscript.

\bibliographystyle{apalike}
\bibliography{facct2022}   

\newpage
\section*{Appendix}
\appendix
\section{The Multi-label approach and Bias}\label{appendix:ml}
\subsection{Why can the multi-label approach  fail?}
We describe the cartoon illustration in Figure \ref{fig:multilabel_fails}. Suppose we have a binary sensitive attribute $\bs$ and three classes. Suppose we have three data records. The first record has the true label probability score $\by=(\frac{1}{2}, \frac{1}{2}, 0)$ as shown in the first row of the matrix $Y$ in the figure. The sensitive attribute of the first record is $\bs=0$, as shown in the first row of $\bs$ in the figure.
The same goes for the second and third records. 

According to the original labels $Y$, $\mathbb{E}[\by|\bs=0]$ is the average of the first and third rows in $Y$, which is $(\frac{1}{4}, \frac{1}{4}, \frac{1}{2})$. On the other hand, $\mathbb{E}[\by|\bs=1]$ is equal to the second row of $Y$, which is $(1, 0, 0)$. So, demographic parity exists in the original labels because $\mathbb{E}[\by|\bs=0]\neq \mathbb{E}[\by|\bs=1]$. Using the definition of multiclass bias in (\ref{eq:multiclass:dp}), the level of bias in this case is $3/4$, where the maximum difference occurs in the first label.

The matrix $\hat Y$, on the other hand, is a transformation of the original labels $Y$ that is unbiased. Note that its second row equals the average of the first and last rows. However, it does not provide valid predictions because its rows are not normalized. This corresponds to the multi-label approach. 

After normalization, bias is re-introduced again in the matrix $\tilde Y$. We have $\mathbb{E}[\tilde{\by}|\bs=0] = (\frac{1}{3}, \frac{1}{6}, \frac{1}{2})$ and $\mathbb{E}[\tilde{\by}|\bs=1] = (\frac{2}{5}, \frac{1}{5}, \frac{2}{5})$. The level of bias according to the definition in (\ref{eq:multiclass:dp}) is 0.1, which indeed decreases but remains, nevertheless, far from the prescribed level. This failure does not only occur in contrived settings. As demonstrated in the synthetic data and real-world applications, the multi-label approach can indeed be sub-optimal in practice.  

\subsection{When does the multi-label approach succeed?}
In the previous section, we showed that the multi-label approach can fail in some settings. However, it seems to perform well in most of the experiments. We present an argument for why this is the case, next. 

Let us consider one subpopulation only for now (e.g. $\textbf{s}=0)$. Suppose we have $n$ examples and $L$ classes. 
Suppose that the labels after the multi-label debiasing step (but before normalization) are encoded in a matrix $H\in\mathbb{R}^{n\times L}$, whose rows give the probability of each class but they are not normalized. One way to normalize the rows is using:
\begin{equation*}
    Z = H - \frac{1}{L}[H\cdot \one - \one]\cdot \one^T
\end{equation*}
This is the normalization that minimizes the Frobenius norm $||H-Z||_F$ as shown in the proof of Proposition \ref{prop:optimality}. Now, let's consider the average predictions across the columns:
\begin{align*}
    \frac{1}{n}\,\one^TZ &= \frac{1}{n}\Big( \one^TH-\frac{1}{L}[\one^TH\one-n]\cdot\one^T\Big)
\end{align*}
Let $Z_k$ be the $k$-th column of $Z$ and similarly for $H$. The change in the average predictions before and after normalization satisfies:
\begin{align*}
    \max_{k\in\{1,2,\ldots,L\}}\Big\{\one^T(Z_k-H_k)\Big\} \le \frac{1}{n}||\one^T(Z-H)||_1 = \frac{|\one^TH\one-n|}{n}
\end{align*}

Now, suppose that $H = Y + E$, where $Y$ is the original matrix of labels (i.e. it is normalized) and $E\in[-1,1]^{n\times L}$ is a random perturbation matrix that has a zero mean. In other words, we consider the case in which debiasing the labels separately behaves like a random perturbation to the original matrix $Y$. Then:
\begin{equation*}
        \max_{k\in\{1,2,\ldots,L\}}\Big\{\one^T(Z_k-H_k)\Big\} \le  \frac{|\one^T(Y+E)\one-n|}{n} = \frac{|\one^TE\one|}{n}.
\end{equation*}
However, $\one^TE\one/n$ is the average of $n$ random variables under our assumption that are bounded in the interval $[-L, +L]$. By the Hoeffding bound, we have:
\begin{equation*}
    p\Big\{\frac{|\one^TE\one|}{n}\ge \epsilon\Big\} \le 2\exp\Big\{-\frac{\epsilon^2n}{2L^2}\Big\}
\end{equation*}

In our case, $H$ is a debiased matrix for a single subpopulation, which means that $\mathbb{E}[\by\,|\,\textbf{s}] = \one^TH/n$ is the same for all groups. By the union bound, the probability that the average predictions do not deviate by more than $\delta$ in any single label is bounded by:
\begin{equation*}
    \sum_{s=1}^R 2\exp\Big\{-\frac{\epsilon^2n_s}{2L^2}\Big\},
\end{equation*}
where $n_s$ is the number of training examples for the group $s$. This suggests that the multilabel approach is likely to work well when all of the groups have a large number of training examples available and if debiasing the labels separately behaves indeed like a random perturbation to the original label matrix. 

\section{Transforming the features: equal-mass binning vs. equal-width}\label{appendix:dpremover}
We use in our implementation of the DP remover algorithm of \citet{feldman2015certifying} \emph{equal-mass} binning, instead of equal-width binning, when estimating the cumulative distribution function (CDF) of each feature. It has been noted that equal-mass binning seems to be a better estimator than equal-width binning \cite{roelofs2020mitigating}. We demonstrate below how equal-width binning using the Freedman–Diaconis rule \cite{fd_rule} can fail at eliminating bias in the DP remover algorithm. 

Suppose that the sensitive attribute $\bs$ is binary and let the distirbution of features $\bx\in\mathbb{R}$ be given by:
\begin{align*}
    p(\bx = x\,|\bs = 0) &= \mathcal{N}(0, \epsilon^2)\\
    p(\bx = x\,|\bs = 1) &= \frac{1}{2}\,\mathcal{N}(0, \epsilon^2) + \frac{1}{2}\,\mathcal{N}(1, \epsilon^2)
\end{align*}
Then, when $\epsilon\ll1$, the interquantile range in the class $\bs=1$ is much larger than $\epsilon$. In the  Freedman–Diaconis rule \cite{fd_rule}, the bin width is proporitonal to the interquantile range. Hence, for any sample size $n$, there exists a constant $\epsilon\ll1$ that is small enough for the entire data to fall into two bins only when $\bs =1$. The CDF function in the latter case is flat everywhere except at $\{0, 1/2, 1\}$. 

On the other hand, the CDF using the same Freedman–Diaconis rule is a continuous function for the class $\bs=0$ (because it comprises of a single Gaussian density). 

Now, consider the \lq\lq median distribution" used in the DP remover algorithm. Let $F_s(\tau)^{-1}$ be the $\tau$ quantile of the features conditioned on $\bs=s$. When, $\bs=1$, the feature value $\bx$ will be mapped to one of three values only:
\begin{equation*}
    \Big\{\frac{F_0(0)^{-1} + F_1(0)^{-1}}{2},\, \frac{F_0(0.5)^{-1} + F_1(0.5)^{-1}}{2},\,\frac{F_0(1)^{-1} + F_1(1)^{-1}}{2}\Big\}
\end{equation*}
When $\bs=0$, on the other hand, a feature value $x$ is mapped to $(F_0(\tau)^{-1} + F_1(\tau)^{-1})/2$, where $\tau=F_0(x)$. In the latter case, the range of the mapping is not restricted to three values only  as in the former case. Hence, the two distributions will be different, and the features are no longer unbiased.

\section{Scaling Effects of the Number of Classes on the Multiclass DP Definition}\label{appendix:scaling}

The definition of demographic parity in (\ref{eq:multiclass:dp}) is computed over the underlying distribution of data. In practice, one measures bias on a \emph{finite} sample. In the finite sample setting, even a perfectly unbiased predictor may appear to be biased according to (\ref{eq:multiclass:dp}) when the number of classes is large due to the nature of the distribution of extreme values.  More precisely, let $L>1$ be the number of classes and suppose that the classifier is unbiased; i.e. $\epsilon=0$ with respect to the data distribution and assume that the central limit theorem holds when estimating the mean outcomes per group $\mathbb{E}_{\bx}[f_k(\bx)\,|\,g(x)]$. Let $\hat \epsilon$ be the \emph{empirical} estimate of bias. Then, a well-known result states that $\hat\epsilon = \Theta(\sqrt{\log L})$ \citep{massart2007concentration}, which increases (albeit slowly) as the number of classes $L$ grows. This is one scaling effect of the number of classes $L$ on the definition of demographic parity. 

Another scaling effect relates to \emph{sparsity}: as the number of classes grows, the predictions $f_k(\bx)$ of a single label $k\in\mathcal{Y}$ become sparser. More precisely, since $\sum_k f_k(\bx)=1$, we have $\mathbb{E}_{\bx}[f_k(\bx)] = 1/L$ on average. Due to this scaling effect, the level of demographic parity usually \emph{decreases} as the number of classes increases, which agrees with the experimental results (see Figure \ref{fig:synth_bias_and_acc}). 

A third scaling effect is class imbalance. Given $L$ classes and $n$ training examples, consider the case in which all classes are equally probable; i.e. $p(\by=k)=1/L$ for all $k\in[L]$. Then, class imbalance can arise even in this setting due to the random sampling of training examples. For instance, if $n_k$ is the number of examples assigned to a class $k$,  $n_k$ has a binomial distribution with probability of success $p=1/L$ and $n$ trials. If $L\gg 1$,  $\mathbb{E}|n_k-n_j|=\Theta(\sqrt{n/L})$ holds in expectation when $k\neq j$. On the other hand, $\mathbb{E}[n_k] = n/L$. Hence, the number of examples $n$ should satisfy $n\gg L$ for classes to be balanced. This is particularly important for demographic parity because the level of bias defined in (\ref{eq:multiclass:dp}) holds uniformly over \emph{all} classes. While the formulation in (\ref{eq:multiclass:dp}) is independent of class imbalance, the estimate of $\epsilon$ might be less reliable if $n_k$ is small (i.e. class $k$ is rare) for some value of $k$.

\section{Proof of Proposition \ref{prop:optimality}}\label{appendix:proof_opt}
First, we note that the optimization problem in (\ref{eq:old_obj}) is convex. The objective function is quadratic on the optimization variable $\hat y$. In addition, the feasible set is convex because the maximum of affine functions is convex while the minimum of affine function is concave \citep{boyd2004convex}. Hence:
\begin{equation*}
    \max_{s\in\mathcal{S}}\,\mathbb{E}_{\bx}[\hat y_k(\bx)\,|\,g(\bx)=s] \,-\,\min_{s\in\mathcal{S}}\,\mathbb{E}_{\bx}[\hat y_k(\bx)\,|\,g(\bx)=s]
\end{equation*}
is a sum of two convex functions on $\hat y$, therefore it is convex. Taking the maximum across $k$ labels retains convexity because the maximum of convex functions is convex \citep{boyd2004convex}. 

Let $\mathcal{C}$ be the set of feasible functions $\hat y:\mathcal{X}\to\mathbb{R}^L$ that satisfy the constraint: $\forall x\in\mathcal{D}:\;0\le \hat y(x)\le 1$ and the bias constraint in (\ref{eq:multiclass:dp}). Then, the optimization problem in (\ref{eq:new_obj}) can be rewritten in the form:
\begin{align}
    &\min_{\hat y\in\mathcal{C}}\;&& \sum_{x\in\mathcal{D}}\,\Big\{(1/2)\,||\hat y(x)||_2^2 \;-\; \hat y(x)^Tf(x) \;+\;  \mathbb{I}_{\mathcal{F}}(z(x))\Big\}.\\
    \nonumber&\text{s.t.}
    &&\forall x\in\mathcal{D}:\;\hat y(x) = z(x)
\end{align}
The augmented Lagrangian using the scaled dual variables \cite{admm_boyd} is:
\begin{equation*}
    L_\tau(\hat y, z, u) = \sum_{x\in\mathcal{D}}\,\Big\{(1/2)\,||\hat y(x)||_2^2 \;-\; \hat y(x)^Tf(x) \;+\;  \mathbb{I}_{\mathcal{F}}(z(x)) \;+\; (\tau/2)\, ||\hat y(x) - z(x) + u(x)||^2\Big\},
\end{equation*}
whose domain on its first argument is the feasible set $\hat y\in\mathcal{C}$. The scaled form of the ADMM updates are \cite{admm_boyd}:
\begin{align*}
    \hat y(x)^{(t+1)} &=\arg\min_{\hat y\in\mathcal{C}}\;\sum_{x\in\mathcal{D}}\,\Big\{(1/2)\,||\hat y(x)||_2^2 \;-\; \hat y(x)^Tf(x)  \;+\; (\tau/2)\, ||\hat y(x) - z^{(t)}(x) + u^{(t)}(x)||^2\Big\}\\
    z(x)^{(t+1)} &= \Pi_\mathcal{F} \big\{\hat y^{(t+1)}(x) + u^{(t)}(x)\big\},\quad\quad
    u(x)^{(t+1)} = u(x)^{(t)} + \hat y(x)^{(t+1)} - z(x)^{(t+1)},
\end{align*}
where $\Pi_{\mathcal{F}}(w(x))$ is the projection into the set $\{v\in\mathbb{R}^L: \,\one^Tv=1\}$. Using Lagrange duality, it can be shown that the solution to the latter projection problem has the closed-form:
    \begin{equation}\label{eq:z}
        z(x)^{(t+1)} = h^{(t+1)}(x) + u^{(t+1)}(x) - \mu(x)\one,
    \end{equation}
    where:
    \begin{equation*}
        \mu(x) = \frac{\one^T(h(x)+u(x))-1}{K}.
    \end{equation*}
Denote $x_i$ for the $i$-th instance in the training dataset. Let $H^{(t)}, Z^{(t)}, U^{(t)} \in \mathbb{R}^{N\times L}$ be matrices, where $N$ is the number of training examples and $L$ is the number of classes, such that $H_{ik}^{(t)} = \hat y_k^{(t)}(x_i)$, $Z_{ik}^{(t)} =  z_k^{(t)}(x_i)$, and $U_{ik}^{(t)} =  u_k^{(t)}(x_i)$. Then, Algorithm \ref{algorithm} implements such ADMM updates in matrix form. Convergence and optimality are guaranteed by ADMM since the optimization problem is convex and contains a strongly-convex term \cite{admm_boyd,nishihara2015general}.

\section{Proof of Proposition \ref{prop:bias}}\label{appendix::proof_bias}
Consider a single subpopulation in $\mathcal{X}_\mathcal{S}$ with $n_0$ training examples. 
We have by the triangle inequality:
\begin{align*}
    \big|\mathbb{E}[f_k(\bx)]-\hat{\mathbb{E}}[h_k(\bx)]\big| \;\le\; \big|\mathbb{E}[f_k(\bx)] - \hat{\mathbb{E}}[f_k(\bx)]\big| \,+\, \big|\hat{\mathbb{E}}[f_k(\bx)]-\hat{\mathbb{E}}[h_k(\bx)]\big|.
\end{align*}
We consider the first term. First, if $f_k\in\mathcal{F}_k$ and $\mathcal{F}_k$ has a Rademacher complexity $d$, then with probability of at least $1-\delta$, the following bounds holds uniformly for all functions $f_k\in\mathcal{F}_k$ \cite{bousquet2003introduction}:
\begin{equation}\label{proof_bias_1}
    \big|\mathbb{E}[f_k(\bx)] - \hat{\mathbb{E}}[f_k(\bx)]\big| \le 2d + \sqrt{\frac{\log\frac{2}{\delta}}{2n_0}}.
\end{equation}
By the union bound, this holds for all the $L$ classes in $\mathcal{Y}$ and $R$ sub-populations in $\mathcal{S}$ with a probability of at least $1-LR\delta$. Therefore, with a probability of at least $1-\delta$, we have:
\begin{equation}\label{eq:proof_bias1}
    \sup_{k\in\mathcal{Y},\,s\in[R]}\;\;\sup_{f_k\in\mathcal{F}_k}\,\big|\mathbb{E}[f_k(\bx)\,|\,g(\bx)=s] - \hat{\mathbb{E}}[f_k(\bx)\,|\,g(\bx)=s]\big| \le 2d + \sqrt{\frac{\log\frac{2LR}{\delta}}{2n_0}}.
\end{equation}
On the other hand:
\begin{align*}
    DP(f) &= \sup_{k\in\mathcal{Y}}\,\Big\{\max_{s\in[R]}\mathbb{E}[f_k(\bx)\,|\,g(\bx)=s]\,-\,\min_{s\in[R]}\mathbb{E}[f_k(\bx)\,|\,g(\bx)=s]\Big\}\\
    &\le \sup_{k\in\mathcal{Y}}\,\Big\{\max_{s\in[R]}\hat{\mathbb{E}}[h_k(\bx)\,|\,g(\bx)=s]\,-\,\min_{s\in[R]}\hat{\mathbb{E}}[h_k(\bx)\,|\,g(\bx)=s]\Big\} \\
    &\quad + 2\,\sup_{k\in\mathcal{Y},\,s\in[R]}\;\;\sup_{f_k\in\mathcal{F}_k}\,\big|\mathbb{E}[f_k(\bx)\,|\,g(\bx)=s] - \hat{\mathbb{E}}[h_k(\bx)\,|\,g(\bx)=s]\big|\\
    &\le \epsilon \,+\, 2\,\sup_{k\in\mathcal{Y},\,s\in[R]}\;\;\sup_{f_k\in\mathcal{F}_k}\,\big|\mathbb{E}[f_k(\bx)\,|\,g(\bx)=s] - \hat{\mathbb{E}}[h_k(\bx)\,|\,g(\bx)=s]\big|
\end{align*}
Here, we used the fact that $h(\bx)$ is debiased using Algorithm \ref{algorithm} so it satisfies demographic parity on the training examples with level $\epsilon$ (by Proposition \ref{prop:optimality}). Therefore, using (\ref{eq:proof_bias1}), we have with a probability of at least $1-\delta$:
\begin{align*}
DP(f)   &\le \epsilon \,+\,2\sup_{k\in\mathcal{Y},\,s\in[R]} \big|\hat{\mathbb{E}}[f_k(\bx)\,|\,g(\bx)=s] - \hat{\mathbb{E}}[h_k(\bx)\,|\,g(\bx)=s]\big| \,+\, 2d\,+\,\sqrt{\frac{\log\frac{2LR}{\delta}}{2n_0}}\\
    &=\epsilon \,+\,2\tau\,+\,2d\,+\,\sqrt{\frac{\log\frac{2LR}{\delta}}{2n_0}}.
\end{align*}

\section{Top-K Results for Adult Income Dataset}\label{app_topk}
See Table \ref{tab:topkadult}. 
\begin{table*}[h]
    \centering
    \scriptsize
    \caption{A summary of the top-2 and top-3 results (99\% confidence intervals) for the five debiasing algorithms on the Adult Income benchmark dataset, where the goal is to predict the education level. See Table \ref{tab:postprocessing_adult_default} for further details about the experiment setup.
    }\vspace{1mm}
    
    \begin{tabular}{ll|c@{\hspace{1.\tabcolsep}}c|c@{\hspace{1.\tabcolsep}}c|c@{\hspace{1.\tabcolsep}}c}
    \toprule
         $p$ &  & \multicolumn{2}{c}{BL} & \multicolumn{2}{c}{DPR} & \multicolumn{2}{c}{OST}   \\
         & & {\sc top-2 [\%]} & {\sc top-3 [\%]} & {\sc top-2 [\%]} & {\sc top-3 [\%]} & {\sc top-2  [\%]} & {\sc top-3 [\%]} \\
         \midrule
0.0&\sc rf&${65.3\pm0.5}$&$77.1\pm0.3$&${63.6\pm0.5}$&$75.6\pm0.3$&${65.3\pm0.5}$&$77.1\pm0.3$\\&\sc knn&${52.7\pm0.4}$&$64.4\pm0.5$&${49.8\pm0.4}$&$62.1\pm0.2$&${52.2\pm0.4}$&$64.0\pm0.5$\\\midrule0.5&\sc rf&${63.4\pm0.4}$&$76.6\pm0.3$&${59.5\pm0.4}$&$72.8\pm0.5$&${55.4\pm0.6}$&$71.6\pm0.4$\\&\sc knn&${51.6\pm0.5}$&$62.5\pm0.5$&${49.2\pm0.5}$&$60.1\pm0.5$&${41.4\pm9.5}$&$53.4\pm11.8$\\\bottomrule
   \end{tabular}

    \begin{tabular}{ll|c@{\hspace{1.\tabcolsep}}c|c@{\hspace{1.\tabcolsep}}c|c@{\hspace{1.\tabcolsep}}c}
    
         $p$ &  & \multicolumn{2}{c}{HTO} & 
         \multicolumn{2}{c}{R2B\textsubscript{0}} &
         \multicolumn{2}{c}{R2B (ours)}  \\
         & & {\sc top-2 [\%]} & {\sc top-3 [\%]} & {\sc top-2 [\%]} & {\sc top-3 [\%]} & {\sc top-2  [\%]} & {\sc top-3 [\%]} \\
         \midrule
0.0&\sc rf&${65.2\pm0.6}$&$76.9\pm0.3$&${65.1\pm0.6}$&$76.8\pm0.3$&${65.0\pm0.6}$&$76.8\pm0.3$\\&\sc knn&${52.1\pm0.4}$&$63.9\pm0.6$&${52.1\pm0.4}$&$63.9\pm0.6$&${51.9\pm0.4}$&$63.7\pm0.4$\\\midrule0.5&\sc rf&${55.1\pm0.6}$&$71.8\pm0.5$&${54.9\pm1.0}$&$71.7\pm0.7$&${55.0\pm0.3}$&$66.0\pm0.4$\\&\sc knn&${44.0\pm0.6}$&$56.9\pm0.5$&${44.0\pm0.5}$&$57.0\pm0.5$&${44.4\pm0.8}$&$57.6\pm0.8$\\\bottomrule
   \end{tabular}
    \label{tab:topkadult}
    \vspace{-2mm}
\end{table*}

\section{Random Forests Classifier Figures on Synthetic Data}\label{appendix:synth_full}
\begin{figure}
    \centering\footnotesize
    Demographic Parity (lower is better)\\[1mm]
    \includegraphics[width=\columnwidth]{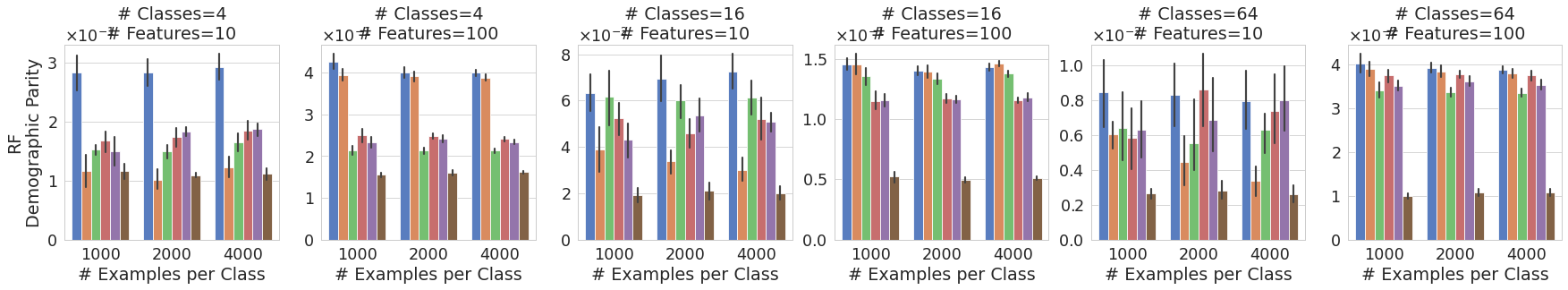}\\
    \includegraphics[width=\columnwidth]{figures/synth_bias_1.png}\\[2mm]
    Prediction Accuracy (higher is better)\\[1mm]
    \includegraphics[width=\columnwidth]{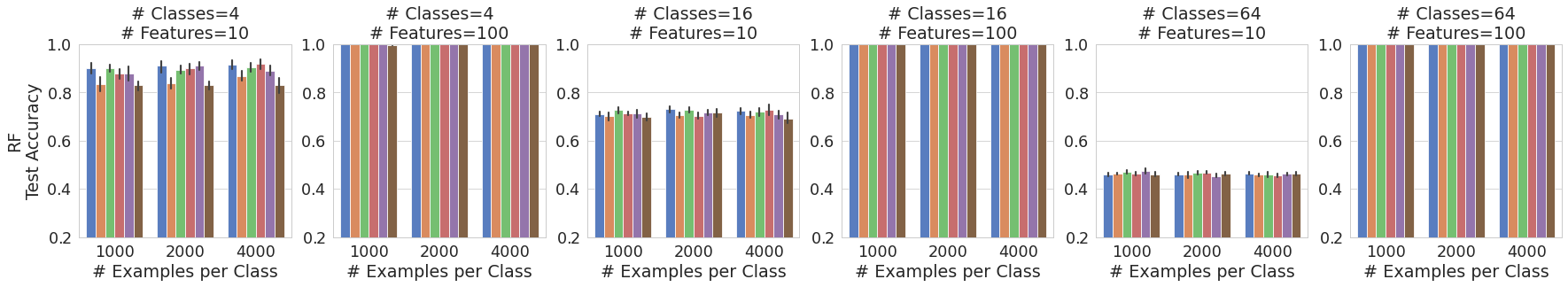}\\
    \includegraphics[width=\columnwidth]{figures/synth_acc_1.png}
    \caption{The top two rows display the multiclass demographic parity as measured using (\ref{eq:multiclass:dp}) for the synthetic dataset with varying numbers of classes, features, and training examples. The first row is for the Random Forest (RF) classifier while the second row is for $k$-NN. Each color represents a debiasing technique. The reduction-to-binary (R2B) method provides a stronger fairness guarantee than the competing methods. In addition, we observe that transforming the features performs poorly when the number of features is large (see the discussion in Section \ref{sect::exp}). The bottom two rows show the prediction accuracy in each case. Note that the drop in accuracy in the case of 64 classes with 10 features is expected because the 64 classes have a large overlap in that setting.}
    \label{fig:synth_bias_and_acc_appendix}
\end{figure}

\begin{figure}
    \centering
    \includegraphics[width=\columnwidth]{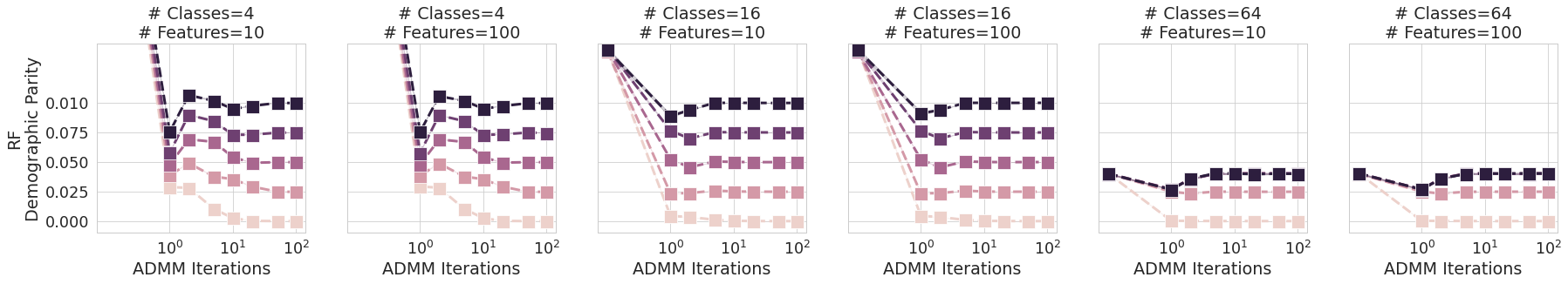}\\
    \includegraphics[width=\columnwidth]{figures/bias_vs_admm_rounds2.png}
    \caption{The level of bias in the \emph{training} data is plotted  as measured by (\ref{eq:multiclass:dp}) vs. the number of ADMM rounds for different levels of bias tolerance $\epsilon\in\{0, 0.025, 0.05, 0.075, 0.10\}$ (see legend). In general, 50-100 rounds of ADMM are sufficient to reach convergence in agreement with prior observations \cite{admm_boyd}. In addition, the level of bias in the training data indeed converges to the minimum of the original data bias level and the prescribed tolerance level $\epsilon$. 
    The number of ADMM rounds in each figure is varied in the set $\{1, 10, 20, 50, 100\}$, where the leftmost point corresponds to original bias in the data.}
    \label{fig:synth_addconvergence_appendix}
\end{figure}

In addition to the $k$-NN figures for the synthetic data experiment, we also provide here similar results using the random forests classifiers. See Figures \ref{fig:synth_bias_and_acc_appendix} and \ref{fig:synth_addconvergence_appendix}.

\section{Training Procedure for COCO Dataset.}\label{appendix:coco_training}
We train a linear classifier with softmax activations and cross entropy loss using the 2,048 pre-logit features of a ResNet50 backbone \citep{he2016deep} pretrained on ImageNet-ILSVRC2012 \citep{deng2009imagenet}. We use SGD with momentum 0.9 for 400 epochs and the following learning rate schedule: $10^{-3}$ (200 epochs), $10^{-4}$ (100 epochs) and $10^{-5}$ (100 epochs).

\section{Dermatology ethics approval and data availability}
\label{app:derm_data_avail}
The images and metadata were de-identified according to Health Insurance Portability and Accountability Act (HIPAA) Safe Harbor prior to transfer to study investigators. The protocol was reviewed by [Anonymous] IRB, which determined that it was exempt from further review under 45 CFR 46. The dermatology data is not available to the public.

\section{Running Time Analysis}
The reduction-to-binary (R2B) algorithm is a lightweight preprocessing method because it operates on the labels only. In Figure \ref{fig:synth_addconvergence_appendix}, we show that ADMM converges quickly in a few rounds in agreement with earlier observations in the literature \citep{admm_boyd}. The time complexity for each single label is $O(|\mathcal{S}|)$ in each round, where $\mathcal{S}$ is the set of demographic groups \citep{alabdulmohsin2021}. Within each group, the required running time can be  controlled according to the desired level of accuracy. For example, using a batch size of 256 and a total of 100K SGD steps, training on a single NVIDIA V100 Tensor Core GPU takes approximately 5s per label in each round of R2B. However, labels can be debiased in parallel in each round of the R2B algorithm.

\end{document}